# Evaluation of Three Vision Based Object Perception Methods for a Mobile Robot

Arnau Ramisa[†], David Aldavert[‡], Shrihari Vasudevan[§],
Ricardo Toledo[‡], Ramon Lopez de Mantaras[†]

[†]*Artificial Intelligence Research Institute (IIIA-CSIC), Campus UAB, E-08193 Bellaterra, Spain, aramisa@iiia.csic.es, mantaras@iiia.csic.es*
[‡]*Computer Vision Center, Campus UAB, E-08193 Bellaterra, Spain, aldavert@cvc.uab.cat, toledo@cvc.uab.cat*
[§]*Australian Centre for Field Robotics, The University of Sydney, NSW 2006, Australia, shrihari.vasudevan@ieee.org*

**Abstract**

This paper addresses object perception applied to mobile robotics. Being able to perceive semantically meaningful objects in unstructured environments is a key capability in order to make robots suitable to perform high-level tasks in home environments. However, finding a solution for this task is daunting: it requires the ability to handle the variability in image formation in a moving camera with tight time constraints. The paper brings to attention some of the issues with applying three state of the art object recognition and detection methods in a mobile robotics scenario, and proposes methods to deal with windowing/segmentation. Thus, this work aims at evaluating the state-of-the-art in object perception in an attempt to develop a lightweight solution for mobile robotics use/research in typical indoor settings.

# 1 Introduction

Currently there is a big push towards semantics and higher level cognitive capabilities in robotics research. One central requirement towards these capabilities is to be able to identify higher level features like objects, doors, etc.

Although impressive results are obtained by modern object recognition and classification methods, still a lightweight object perception method is lacking. Furthermore, the system should be able to learn new objects in an easy, and preferably automatic, way.

For example, in [1], the authors investigate underlying representations of spatial cognition for autonomous robots. Although not specifically addressed in that work, object perception is an essential component that the authors reported to be the most limiting factor.

Although different modalities of perception (e.g. laser range-finder, color camera, haptics) can be used, in this work we focus on passive vision, as it is interesting for several reasons, like an affordable cost, compatibility with human environments or richness of perceived information.

Another example of it can be found looking at the poor results achieved in the detection test of the Pascal 2007 chall. Recently several methods have been quite successful in particular instances of the problem, such as detecting frontal faces or cars, or in datasets that concentrate on a particular issue (e.g. classification in the Caltech-101 [2] dataset). However in more challenging datasets like the detection competition of the Pascal VOC 2007 [3] the methods presented achieved a lower average precision. This low performance is not surprising, since object recognition in real scenes is one of the most challenging problems in computer vision [4]. The visual appearance of objects can change enormously due to different viewpoints, occlusions, illumination variations or sensor noise. Furthermore, objects are not presented alone to the vision system, but they are immersed in an environment with other elements, which clutter the scene and make recognition more complicated.

In a mobile robotics scenario a new challenge is added to the list: computational complexity. In a dynamic world, information about the objects in the scene can become obsolete even before it is ready to be used if the recognition algorithm is not fast enough.

In the present work our intent is to survey some well established object recognition systems, comment on its applicability to robotics and evaluate them on a mobile robotics scenario. The selected methods are the SIFT object recognition algorithm[11], the Bag of Features[14], and the Viola and Jones boosted cascade of classifiers[17], and they were chosen taking into consideration issues relevant to our objective, for example its ability to detect at the same time they recognize, its speed or scalability and the difficulty of training the system. From the obtained results we extract our conclusions and propose several modifications to improve the performance of the methods. Namely, we propose improvements to increase the precision of the SIFT object recognition method, and a segmentation approach to make the Bag of Features method suitable for detection in interactive time. We also benchmark the proposed methods against the typi-cally used Viola and Jones classifier. Finally, we perform extensive tests with the selected methods in our publicly available dataset[1] to assess their performance in a mobile robotics setting.

The three methods are fundamentally different in that they address recognition, classification and detection (three core problems of visual perception), but still can be tailored to the other objectives too. We compare and benchmark these three successful vision approaches towards use in real mobile robotics experiments, providing an useful guide for roboticists who need to enable their robots with object recognition capabilities. The selected algorithms are evaluated under different issues, namely:

- **Detection**: Having the ability to detect where in the image is located the object. In most situations, large portions of the image are occupied by background objects that introduce unwanted information which may confuse the object recognition method.

- **Classification**: A highly desirable capability for an object detection method is to be able to generalize and recognize previously unseen instances of a particular class.

- **Occlusions**: Usually a clear shot of the object to recognize will not be available to the robot. An object recognition method must be able to deal with only partial information of the object.

- **Texture**: Objects with a rich texture are typically easier to recognize than those only defined by its shape and color. We want to evaluate the behavior of each method with both types of objects.

- **Repetitive patterns**: Some objects, such as a chessboard, present repetitive patterns that cause problems in methods that have a data association stage.

- **Training set resolution**: Large images generate more features at different scales (specially for smaller ones) that are undoubtedly useful for object recognition. However, if training images have a resolution much higher than test images, descriptors may become too different.

- **Training set size**: Most methods can benefit from a larger and better annotated training set. However, building such a dataset is time consuming. We want to assess which is the least amount of training information that each method requires to obtain its best results.

- **Run-Time**: One of the most important limitations of the scenario we are considering is the computation time. We want to measure the frame-rate at which comparable implementations of each method can work.

---

[1]Available for download at `http://www.iiia.csic.es/~aramisa/iiia30.html`

- **Detection accuracy**: Computing accurately the location of the object can significantly benefit other tasks such as grasping or navigation. We are interested in quantifying the precision of the object detection in the object recognition algorithm according to the ground truth.

Although different parts of object recognition methods (e.g. feature detectors and descriptors, machine learning methods) have been extensively compared in the literature, to our knowledge there is no work that compares the performance of complete object recognition methods in a practically hard application like mobile robotics.

Probably the work most related with ours is the one of [5], where four methods (SIFT and KPCA+SVM with texture and color features) were combined in an object recognition/classification task for human-robot interaction. The appropriate method for each class of object was chosen automatically from the nine combinations of task/method/features available, and models of the learned objects were improved during interaction with the user (pictured as a handicapped person in the paper). This work was, however, more focused on building a working object classification method suitable for the particular task of human-robot interaction with feedback from the human user, and not in evaluating each particular method in a standardized way. Furthermore, no quantitative results were reported for the experiments with the robot.

Mikolajczyk et al. [6, 7] do a comprehensive comparison of interest region detectors and descriptors in the context of keypoint matching. Although this works are undoubtedly related with the one presented here, the objectives of the comparison are notably different: while Mikolajczyk et al. measured the repeatability of the region detectors and the matching precision of the region descriptors, here we focus on the performance of three well-known object recognition methods in the very specific setting of mobile robotics.

The rest of the paper is divided as follows: First, Table 1 shows the conclusions reached in this work regarding the applicability of the evaluated methods in the mobile robot vision domain. Next, in Section 2 comes an overview of the datasets used in our experimentation. In Sections 3 to 5 the different object recognition algorithms are briefly described and the experiments done to arrive to the conclusions for each presented. Finally, in Section 6, the conclusions of the work are presented and continuation lines proposed.

|  | SIFT | Vocabulary Tree | Cascade of Simple Classifiers |
|---|---|---|---|
| Detection | Can detect objects under in-plane rotation, scale changes and small out-of-plane rotations | Must be complemented with a sliding windows approach, a segmentation algorithm or an interest operator | Is able to determine the most probable bounding box of the object |
| Pose Estimation | Up to an affine transformation | presence/absence only | presence/absence only |
| Classification (intra-class variation and generalization) | No | Yes | Yes |
| Occlusions | Tolerates it as long as at least 3 points can be reliably matched (depends on ammount of texture) | Showed good tolerance to occlusions | Low tolerance to occlusions |
| Repetitive patterns | No | Yes | Yes |
| Minimum training set size | One image | Tens of images | Hundreds or thousands of images |
| Training set resolution | VGA resolution is sufficient | Benefits from higher resolution of training data | VGA resolution is sufficient |
| Run-Time | Less than a second per image | two seconds per image with a segmentation algorithm included | Less than a second per image |

Table 1: Qualitative summary of results found in our experiments.

## 2 Datasets and Performance Metrics

In order to evaluate the methods in a realistic mobile robots setting, we have created the IIIA30 dataset[2], that consists of three sequences of different length acquired by our mobile robot while navigating at approximately 50 cm/s in a laboratory type environment and approximately twenty good quality images for training taken with a standard digital camera. The camera mounted in the robot is a Sony DFW-VL500 and the image size is 640x480 pixels. In Figure 1 the robotic platform used can be seen. The environment has not been modified in any way and the object instances in the test images are affected by lightning changes, blur caused by the motion of the robot, occlusion and large viewpoint and scale changes.

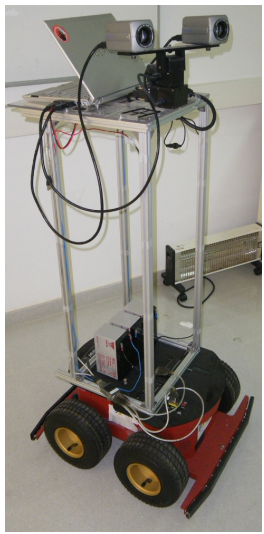

Figure 1: Robotic platform used in the experiments.

We have considered a total of 30 categories (29 objects and background) that appear in the sequences. The objects have been selected to cover a wide range of characteristics: some are textured and flat, like the posters, while others are textureless and only defined by its shape. Figure 2.a shows the training images for all the object categories, and 2.b shows some cropped object instances from the test images. Each occurrence of an object in the video sequences has been manually annotated in each frame to construct the ground truth, along with its particular image characteristics (e.g. blurred, occluded...).

In order to evaluate the performance of the different methods we used several standard metrics that are briefly explained in the following lines. Precision is defined as the ratio of true positives among all the positively labeled examples,

---

[2]http://www.iiia.csic.es/~aramisa/iiia30.html

and reflects how accurate our classifier is.

$$Pre = \frac{TruePositives}{FalsePositives + TruePositives} \tag{1}$$

Recall measures the percentage of true positives that our classifier has been able to label as such. Namely,

$$Rec = \frac{TruePositives}{FalseNegatives + TruePositives} \tag{2}$$

Since it is equally important to perform well in both metrics, we also considered the F-Measure metric:

$$f - measure = \frac{2 \cdot Precision \cdot Recall}{Precision + Recall} \tag{3}$$

This measure assigns a single score to an operating point of our classifier weighting equally precision and recall, and is also known as $f_1 - measure$ or balanced $f - score$. If the costs of a false positive and a false negative are asymetric, the general f-measure can be used by adjusting the $\beta$ parameter:

$$f_g - measure = \frac{(1 + \beta^2) \cdot Precision \cdot Recall}{\beta^2 \cdot Precision + Recall} \tag{4}$$

In the object detection experiments, we have used the Pascal VOC object detection criterion [3] to determine if a given detection is a false or a true positive. In brief, to consider an object as a true positive, the bounding boxes of the ground truth and the detected instance must have a ratio of overlap equal or greater than 50% according to the following equation:

$$\frac{BB_{gt} \cap BB_{detected}}{BB_{gt} \cup BB_{detected}} \geq 0.5 \tag{5}$$

where $BB_{gt}$ and $BB_{detected}$ stand for the ground truth and detected object bounding box respectively. For objects marked as occluded only the visible part has been annotated in the ground truth, but the SIFT object recognition method will still try to adjust the detection bounding box for the whole object based only in the visible part. Since the type of annotation is not compatible with the output of the SIFT algorithm, for the case of objects marked as occluded, we have modified the above formula in the following way:

$$\frac{BB_{gt} \cap BB_{detected}}{BB_{gt}} \geq 0.5 \tag{6}$$

As can be seen in the previous equation, it is only required that the detected object bounding box overlaps 50% of the ground truth bounding box.

Apart from the IIIA30 dataset, in order to test and adjust the parameters of the Vocabulary Tree object recognition method, we have used two pre-segmented image databases:

- ASL: The ASL recognition dataset[3] consists of nine household objects from the Autonomous Systems Lab of the ETHZ [8]. It consists of around 20 training images per object from several viewpoints and 36 unsegmented test images with several instances of the objects, some of them with illumination changes or partial occlusions. The training images have been taken with a standard digital camera at a resolution of 2 megapixels, while the test images have been acquired with a STHMDCS2VAR/C stereo head by Videre design at the maximum possible resolution (1.2 megapixels). A segmented version of the training object instances has also been used in some experiments, and is referred as *segmented ASL*. Some images of the segmented version can be seen in Figure 2.

- Caltech10: This is a subset of the Caltech 101 dataset [9] , widely used in computer vision literature. We have taken 100 random images of the ten most populated object categories, namely: planes (lateral), bonsais, chandeliers, faces (frontal), pianos, tortoises, sails, leopards, motorbikes and clocks as seen in Figure 4. Training and testing subsets are determined randomly in each test. Experiments with this dataset have been done following the setup of [10]: 30 random training images and the rest for testing.

---

[3] http://www.iiia.csic.es/~aramisa/iiia30.html

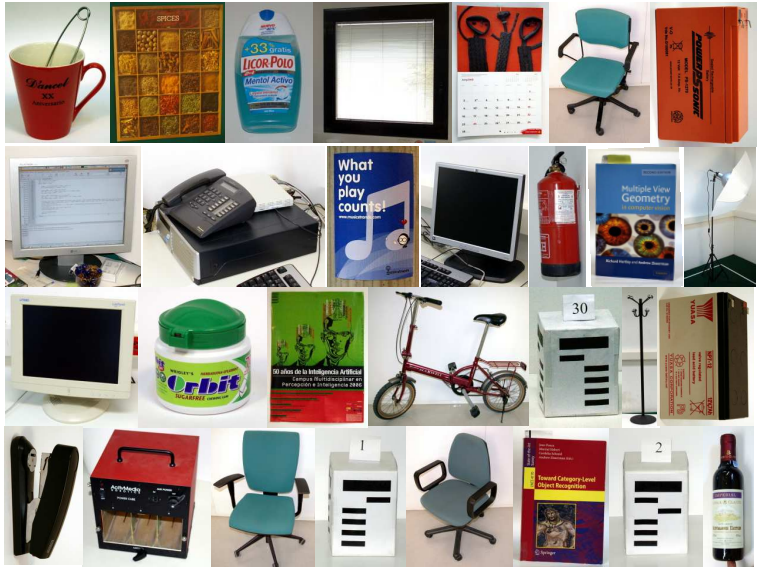

(a)

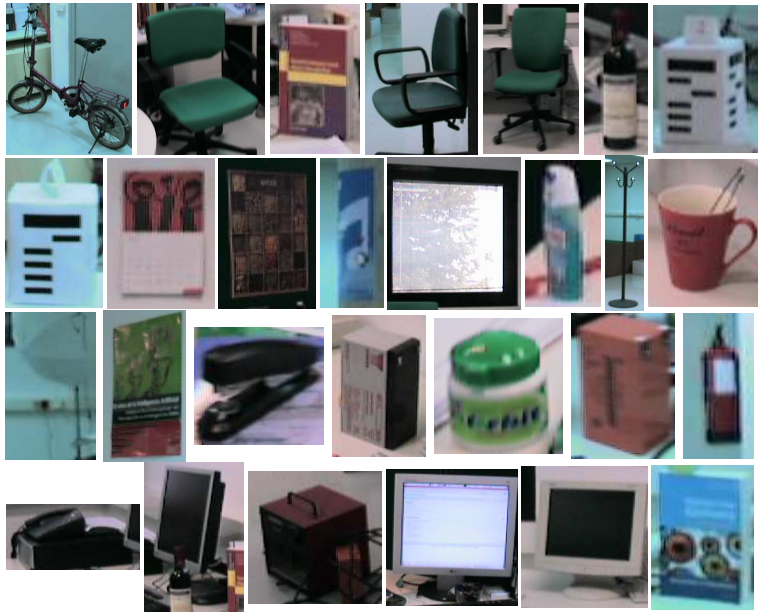

(b)

Figure 2: (a) Training images for the IIIA30 dataset. (b) Cropped instances of objects from the test images.

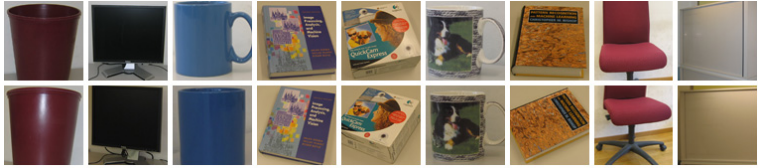

(a)

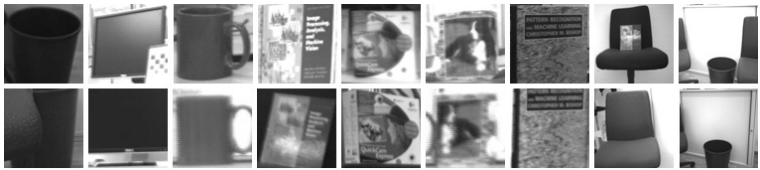

(b)

Figure 3: Segmented ASL dataset images. (a) Training. (b) Testing.

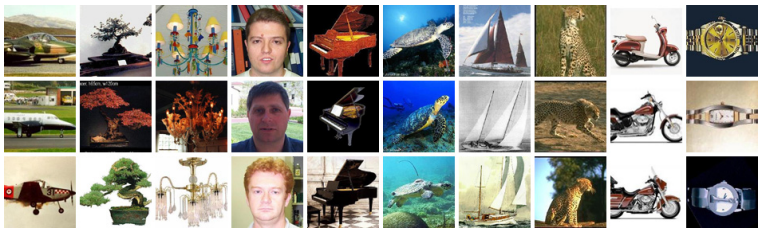

Figure 4: Images from Caltech10 dataset.

## 3 Lowe's SIFT

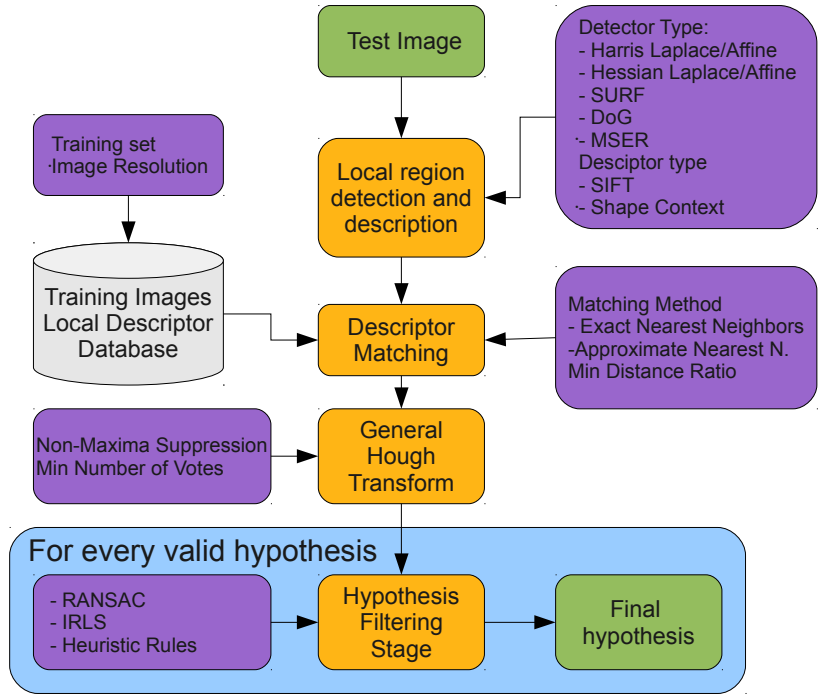

Figure 5: Diagram of the Lowe's SIFT method with all the tests performed shown as purple boxes, Orange ones refer to steps of the method and green to input/output of the algorithm.

Lowe's SIFT object recognition approach is a view-centered object detection and recognition system with some interesting characteristics for mobile robots, most significant of which is the ability to detect and recognize objects in an unsegmented image. Another interesting feature is the Best-Bin-First algorithm used for approximated fast matching, which reduces the search time by two orders of magnitude for a database of 100,000 keypoints for a 5% loss in the number of correct matches [11]. Follows a brief outline of the algorithm.

The first stage of the approach consists on matching individually the SIFT descriptors of the features detected in a test image to the ones stored in the object database using the Euclidean distance. As a way to reject false correspondences, only those query descriptors for which the best match is isolated from the second best and the rest of database descriptors are retained. In Figure 6, the matching features between a test and model images can be seen. The presence of some outliers (incorrect pairings of query and database features) can also be observed.

Once a set of matches is found, the Generalized Hough Transform is used to cluster each match of every database image depending on its particular transformation (translation, rotation and scale change). Although imprecise, this step generates a number of initial coherent hypotheses and removes a notable portion of the outliers that could potentially confuse more precise but also more sensitive methods. All clusters with at least three matches for a particular training object are accepted, and fed to the next stage: the Least Squares method, used to improve the estimation of the affine transformation between the model and the test images.

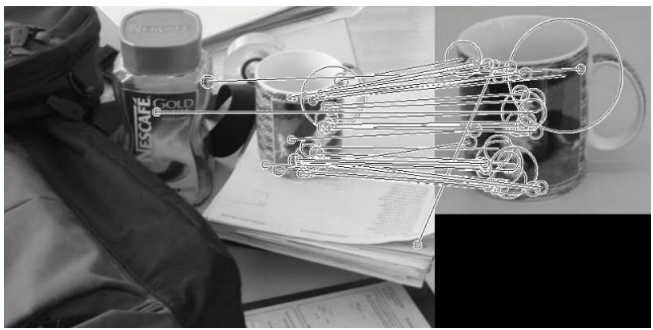

Figure 6: Matching stage in the SIFT object recognition method.

This approach has been modified in several ways in our experiments: The least squares method has a 0% breakdown point (i.e. any false correspondence will make the model fitting method fail or give sub-optimal results), which is a rather unfeasible restriction since we have found it is normal to still have some false matches in a given hypothesis after the Hough Transform.

To alleviate this limitation, instead of the least squares, we have used the Iteratively Reweighted Least Squares (IRLS), which we have found to perform well in practice at a reasonable speed. Furthermore we have evaluated the RANdom SAmple Consensus (RANSAC), another well-known model fitting algorithm, to substitute or complement the IRLS. The RANSAC algorithm iteratively tests the support of models estimated using minimal subsets of points randomly sampled from the input data. Finally, we have incorporated some domain knowledge by defining several heuristic rules on the parameters of the estimated affine transformation to reject those clearly beyond plausibility. Namely:

- Hypotheses with object centers that are too close.

- Hypotheses that have a ratio between the x and y scales below a threshold.

Figure 5 shows an overview of our implementation of the SIFT object recognition algorithm steps.

For evaluating the method, one image per category from the training image set is used. As there are several parameters to adjust in this method, we used the first sequence of the IIIA30 dataset (IIIA30-1) as test data to perform an extensive cross-validation over detector and descriptor type, training image size, matching method, distance ratio to the second nearest neighbor for rejecting matches, non-maxima suppression and minimum number of votes in the Hough Transform and hypothesis verification and refinement methods.

Since this study is too extensive to be included here, details are provided online for the interested reader[4]. Follows a brief summary of the most relevant results obtained with the corss-validation.

In this section the results of cross-validation tests conducted using sequence 1 of the IIIA30 dataset (IIIA30-1) with the different parameter combinations considered are described. Taking into account all combinations, the best recall obtained has been 0.45 with the Hessian Laplace detector and the less restrictive settings possible. However this configuration suffered from a really low precision, just 0.03.

The best precision score has been 0.94, and has been obtained also with the Hessian Laplace detector, with a restrictive distance ratio to accept matches: 0.5. The recall of this combination was 0.14. The same precision value but with lower recall has been obtained with the SURF and Hessian Affine detectors.

Looking at the combinations that had a best balance between recall and precision (best f—measure), the top performing ones obtained 0.39 also with the Hessian Laplace detector (0.29 recall and 0.63 precision). However, even though approximate nearest neighbors is used, each image takes around 2 seconds to be processed.

Given the objectives of this work, the most relevant way to analyze the results consists in prioritizing the time component and select the fastest parameter settings. As a runtime greater than one second is not acceptable for our purposes, the combinations that improved the f—measure with respect to faster combinations for those close to one second for image have been selected as interesting. Table 2 shows the parameters of the chosen combinations. For more information on the experiments conducted please refer to the technical report[3].

Once the parameter combinations that best suited our purposes were found, we evaluated them in all the test sequences.

## 3.1 Evaluation of Selected Configurations

This section presents the results obtained applying the parameter combinations previously selected to all the sequences in the dataset. In general all possible combinations of parameters performed better in well textured and flat objects, like the books or posters. For example the *Hartley book* or the *calendar* had an average recall across the six configurations (see Table 2 for the configuration parameters) of 0.78 and 0.54 respectively. This is not surprising as the SIFT de-scriptor assumes local planarity, and depth discontinuities can severely degrade descriptor similarity. On average, textured objects achieved a recall of 0.53 and a precision 0.79 across all sequences. Objects only defined by shape and color were in general harder or even impossible to detect, as can be seen[5] in Table 3. Recall for this type of objects was only 0.05 on average. Configuration 6, that used the Hessian Laplace detector, exhibited a notably better performance for some objects of this type thanks to its higher number of detected regions. For example the *chair* obtained a recall of 0.54, or the *rack* that obtained a 0.77 recall using this feature detector. Finally, and somewhat surprisingly, objects with a repetitive texture such as the *landmark cubes* (see Figure 2) had a quite good recall of 0.46 on average. Furthermore, the result becomes even better if we take into consideration that besides the self-similarity, all three *landmark cubes* were also similar to one another.

---

[4]http://www.iiia.csic.es/~aramisa/datasets/iiia30.html

| Method | Distance Ratio | Detector | Min. Matches | HT Method | RANSAC | Approx-NN | IRLS | Heuristics | Time (sec) | Recall | Precision | F-Measure |
|---|---|---|---|---|---|---|---|---|---|---|---|---|
| Config 1 | 0.8 | SURF | 5 | NMS | No | Yes | Yes | No | 0.37 | 0.15 | 0.51 | 0.23 |
| Config 2 | 0.8 | SURF | 3 | NMS | Yes | Yes | Yes | Yes | 0.42 | 0.14 | 0.87 | 0.24 |
| Config 3 | 0.8 | DoG | 10 | NMS | No | Yes | Yes | No | 0.52 | 0.17 | 0.47 | 0.25 |
| Config 4 | 0.8 | DoG | 10 | NMS | Yes | Yes | Yes | Yes | 0.55 | 0.17 | 0.9 | 0.28 |
| Config 5 | 0.8 | DoG | 5 | NMS | Yes | Yes | Yes | Yes | 0.60 | 0.19 | 0.87 | 0.31 |
| Config 6 | 0.8 | HesLap | 10 | NMS | Yes | Yes | Yes | Yes | 2.03 | 0.28 | 0.64 | 0.39 |

Table 2: Detailed configuration parameters and results for the six representative configurations in increasing time order. They have been chosen for providing the best results in a sufficiently short time.

| Object | Config 1 | | Config 2 | | Config 3 | | Config 4 | | Config 5 | | Config 6 | |
|---|---|---|---|---|---|---|---|---|---|---|---|---|
| | Rec | Pre | Rec | Pre | Rec | Pre | Rec | Pre | Rec | Pre | Rec | Pre |
| Grey battery | 0 | 0 | 0 | 0 | 0 | 0 | 0 | 0 | 0 | 0 | 0 | 0 |
| Bicycle | **0.54** | 0.52 | 0.52 | **1.00** | 0.33 | 0.52 | 0.36 | 0.89 | 0.38 | 0.90 | 0.33 | 0.62 |
| Hartley book | 0.58 | **0.93** | 0.58 | **0.93** | 0.86 | 0.77 | 0.88 | 0.88 | **0.95** | 0.85 | 0.81 | 0.73 |
| Calendar | 0.44 | 0.65 | 0.35 | **0.86** | 0.56 | 0.66 | 0.56 | 0.79 | 0.56 | 0.79 | **0.79** | 0.71 |
| Chair 1 | 0.03 | 0.08 | 0.02 | 0.33 | 0 | 0 | 0 | 0 | 0.01 | 1.00 | **0.54** | **1.00** |
| Charger | 0.03 | 0.20 | 0.03 | **0.50** | 0 | 0 | 0 | 0 | 0 | 0 | **0.18** | 0.14 |
| Cube 2 | 0.62 | 0.28 | 0.67 | **0.67** | 0.71 | 0.11 | **0.76** | 0.59 | 0.76 | 0.55 | 0.52 | 0.38 |
| Monitor 3 | 0 | 0 | 0 | 0 | 0 | 0 | 0 | 0 | 0 | 0 | **0.02** | **0.33** |
| Poster spices | 0.38 | 0.77 | 0.42 | **0.94** | 0.54 | 0.79 | 0.53 | 0.87 | **0.58** | 0.87 | 0.56 | 0.92 |
| Rack | 0.26 | 0.59 | 0.26 | **1.00** | 0.10 | 0.80 | 0.10 | 1.00 | 0.23 | 1.00 | **0.77** | 0.79 |

Table 3: Object-wise recall and precision for all combinations.

Regarding the image quality parameters (see Table 4), all combinations behaved in a similar manner: the best recall, as expected, was obtained by images not affected by blur, occlusions or strong illumination changes. From the differ-ent disturbances, what was tolerated best was occlusion, followed by blur and then by illumination. Combinations of problems also had a demolishing effect in the method performance as seen in the last three rows of Table 4, being the worst case the combination of *blur* and *illumination* that had 0 recall. Object instance size (for objects with a bounding box defining an area bigger than 5000 pixels) did not seem to have such an impact in performance as image quality has. The performance with objects of smaller area has not yet been rigorously analyzed and is left for future work. As can be seen in the results, RANSAC and the heuristics significantly improved precision without affecting recall.

---
[5]For space reasons, only part of the Table was included. The full Table can be found in http://www.iiia.csic.es/~aramisa/datasets/iiia30_results/results.html

| Object | Config 1 | Config 2 | Config 3 | Config 4 | Config 5 | Config 6 |
|---|---|---|---|---|---|---|
| Normal | 0.26 | 0.25 | 0.26 | 0.28 | 0.3 | 0.33 |
| Blur | 0.1 | 0.1 | 0.16 | 0.15 | 0.18 | 0.25 |
| Occluded | 0.16 | 0.14 | 0.14 | 0.12 | 0.14 | 0.34 |
| Illumination | 0 | 0 | 0.06 | 0.06 | 0.06 | 0.06 |
| Blur+Occl | 0.06 | 0.04 | 0.08 | 0.06 | 0.09 | 0.14 |
| Occl+Illum | 0.08 | 0.08 | 0.08 | 0.08 | 0.08 | 0.06 |
| Blur+Illum | 0 | 0 | 0 | 0 | 0 | 0 |

Table 4: Recall depending on image characteristics. *Normal* stands for object instances with good image quality and *blur* for blurred images due to motion, *illumination* indicates that the object instance is in a highlight or shadow and therefore has low contrast. Finally the last three rows indicate that the object instance suffers from two different problems at the same time.

Finally, we have validated the detection accuracy by the ratio of overlap between the ground truth bounding box and the detected object instance as calculated in Equation 5. As can be seen in Figure 7, on average 70% of true positives have a ratio of overlap greater than to 80%, regardless of the parameter combination. Furthermore, we found no appreciable advantage on detection accuracy for any object type or viewing conditions, although a more in-depth analysis of this should be addressed in future work.

As a means to provide a context to the results obtained with the six selected configurations (i.e. how good are they with respect to what can be obtained without taking into account the execution time), we compare them to the best overall recall and precision values obtained with the SIFT object recognition method. Table 5 displays the averaged precision and recall values of the four configurations that obtained the overall best recall and the four that obtained the overall best precision, as well as the six selected configurations. As can be seen in the table, the attained recall in the selected configurations was 20% lower than the maximum possible, independently of the type of objects. Precision is more affected by the amount of texture, and differences with respect to the top performing configurations ranged from 17% to 38%.

## 3.2 Discussion

Experiments show that, using the SIFT object recognition approach with the proposed modifications, it is possible to precisely detect, considering all image degradations, around 60% of well-textured object instances with a precision close to 0.9 in our challenging dataset at approximately one frame per second in $640 \times 480$ pixel images with our not fully optimized implementation. Even detectors known to sacrifice repeatability (probability of finding the same feature region in slightly different viewing conditions) for speed such as the SURF obtain reasonable results. Performance degrades for objects with repetitive textures or no texture at all. Regarding image disturbances, the approach resisted occlusions well, since the SIFT object recognition method is able to estimate a reliable transformation (as long as a minimum number of correct matches is found, three by default), but not so well blur due to motion or deficient illumination.

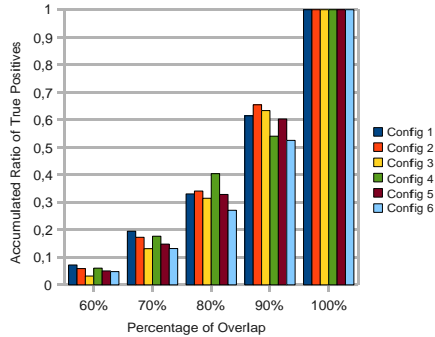

Figure 7: Accumulated frequencies for ratio of overlap between the ground truth bounding box and the detected bounding box for correctly found objects (true positives). An object is considered correctly detected if the ratio of overlap between the bounding boxes computed with equation 5 is 50% or more.

The step of the algorithm that takes most of the processing time is the descriptor matching, as it has a complexity of $O(N \cdot M \cdot D)$ comparisons, where $N$ is the number of features in the new test image, $M$ is the number of features in the training dataset and $D$ is the dimension of the descriptor vector. Approximate matching strategies, such as the one by [12] used in this work, make the SIFT object recognition method suitable for robotic application by largely reducing its computational cost. In our experiments we experienced only a 0.01 loss in the f—measure for an up to 35 times speed-up. Furthermore, an implementation tailored to performance should be able to achieve even faster rates. A drawback of the SIFT object recognition method is that it is not robust to viewpoint change. It would be interesting to evaluate how enhancing the method with 3D view clustering as described in [13] affects the results, as it should introduce robustness to this type of transformation.

|  | Best Recall | | Best Precision | | Selected Config. | |
|---|---|---|---|---|---|---|
|  | mean | std | mean | std | mean | std |
| Repetitively textured objects | | | | | | |
| Recall | 0.65 | 0.09 | 0.16 | 0.01 | 0.46 | 0.05 |
| Precision | 0.02 | 0.01 | 0.75 | 0.15 | 0.43 | 0.24 |
| Textured objects | | | | | | |
| Recall | 0.70 | 0.03 | 0.28 | 0.03 | 0.53 | 0.10 |
| Precision | 0.05 | 0.02 | 0.96 | 0.02 | 0.79 | 0.09 |
| Not textured objects | | | | | | |
| Recall | 0.21 | 0.01 | 0.01 | 0.01 | 0.05 | 0.04 |
| Precision | 0.03 | 0.01 | 0.62 | 0.32 | 0.24 | 0.21 |

Table 5: Average recall and precision of the configurations that where selected for having the best values according to these two measures in the last section. Also average results among the six selected configurations are shown for comparison. Standard deviation is provided to illustrate scatter between the selected configurations. Objects are grouped in the three "level of texture" categories in the following way: the three cubes form the repetitively textured category, the two books, the calendar and the three posters form the textured category, and the rest fall into the non textured category.

## 4 Vocabulary Tree Method

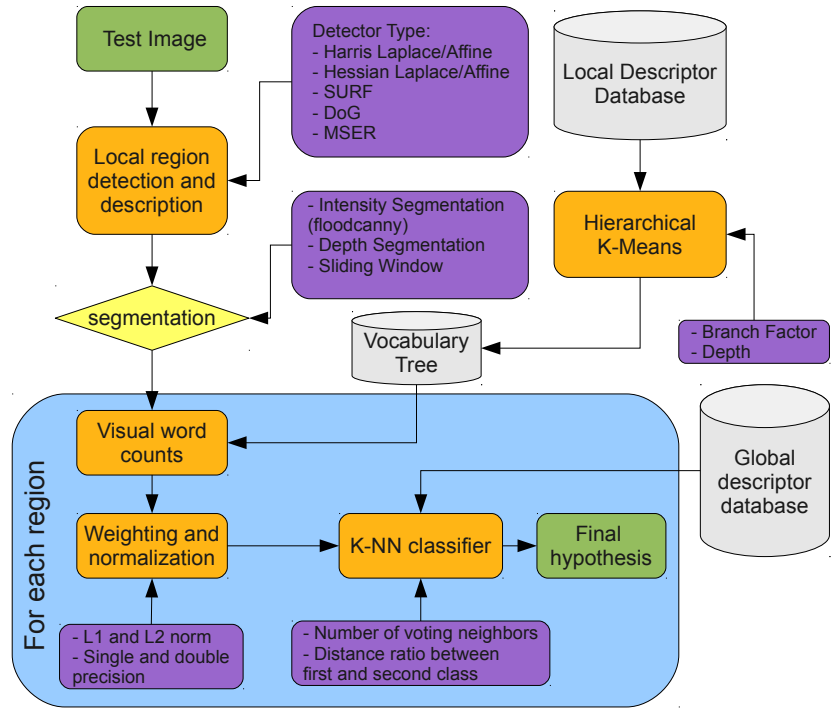

Figure 8: Diagram of the Vocabulary Tree method. Modifications to the original algorithm have yellow background and tests performed are shown as purple boxes. As before, orange boxes refer to steps of the method and green to input/output of the algorithm.

The Vocabulary Tree approach [14] to object classification is based on the bag of words document retrieval methods, that represent the subject of a document by the frequency in which certain words appear in the text. This technique has been adapted to visual object classification substituting the words with local descriptors such as SIFT computed on image features [15, 16].

Although recently many approaches have been proposed following the *bag of words* model, we have selected this particular one because scalability to large numbers of objects in a computationally efficient way is addressed, which is a key feature in mobile robotics. Figure 8 shows the main steps of the [14] algorithm. First the local feature descriptors are extracted from a test image, and a visual vocabulary is used to quantize those features into *visual words*.

A hierarchical vocabulary tree is used instead of a linear dictionary, as it allows to code a larger number of visual features and simultaneously reduce the look-up time to logarithmic in the number of leaves. The vocabulary tree is built using hierarchical k-means clustering, where the parameter $k$ defines the branch factor of the tree instead of the final number of clusters like in the flat (standard) k-means. On the negative side, using such hierarchical dictionaries causes aliasing in cluster space that can reduce the performance of the approach.

Then, the visual words are weighted in accordance to its discriminative power with the Term Frequency-Inverse Document Frequency (TF-IDF) scheme to improve retrieval performance. Let $n_i$ be the number of descriptors corresponding to the codeword $i$ found in the query image and $m_i$ the number of descriptors corresponding to the same codeword for a given training image, and let $q$ and $d$ be the histogram signatures of the query and database images, then the histogram bins $q_i$ and $d_i$ can be defined as:

$$q_i = n_i \omega_i$$
$$d_i = m_i \omega_i \quad (7)$$

where $\omega_i$ is the weight assigned to node $i$. A measure based in entropy is used to define the weights:

$$\omega_i = ln(\frac{N}{N_i}), \quad (8)$$

where $N$ is the number of images in the database, and $N_i$ is the number of images in the database with at least one descriptor vector path through node $i$. Since signatures will be normalized before comparison, the resulting schema is the term frequency-inverse document frequency.

To compare a new query image with a database image, the following score function is used:

$$s(q,d) = \|\frac{q}{\|q\|} - \frac{d}{\|d\|}\| \quad (9)$$

The normalization can be in any desired norm, but the L1-norm (also known as the "Manhattan" distance) was found to perform better both by [14] and in our experiments. The class of the object in the query image is determined as the dominant one in the $k$ nearest neighbors from the database images.

The second speed-up proposed by Nister and Stewenius consists on using *inverted files* to organize the database of training images. In an inverted files structure each leaf node contains the ID number of the images whose signature value for this particular leaf is not zero. To take advantage of this representation, and assuming that the signatures have been previously normalized, the previous equation can be simplified making the distance computation only dependent on the nonzero elements both in the query and database vectors. With this distance formulation one can use the inverted files and, for each node, accumulate to the sum only for the training signatures that have non-zero value. If signatures are normalized using the L2 norm (i.e. the Euclidean distance), the distance computation can be simplified further to:

$$||q - d||_2^2 = 2 - 2 \sum_{i|q_i \neq 0, d_i \neq 0} q_i d_i \qquad (10)$$

and since we are primarily interested in the ranking of the distances, we can simply accumulate the products and sort the results of the different images in descending order.

The main drawback of the Vocabulary Tree method is that it needs at least a rough segmentation of the object to be recognized. The most straightforward solution to overcome this limitation is to divide the input image using a grid of fixed overlapping regions and process each region independently. Alternatively, we propose a fast segmentation algorithm to generate a set of meaningful regions that can later be recognized with the vocabulary tree method.

The first option has the advantage of simplicity and universality: Results do not depend on a particular method or set of segmentation parameters, but just on the positions and shapes of the windows evaluated. However a square or rectangular window usually does not fit correctly the shape of the object we want to detect and, in consequence, background information is introduced. Furthermore, if we want to exhaustively search the image, in the order of $O(n^4)$ overlapping windows will have to be defined, where $n$ is the number of pixels of the image. This will be extremely time-consuming, and also fusing the classification output of the different windows into meaningful hypotheses is a non-trivial task. One way that could theoretically speed-up the sliding window process is using integral images [17]. This strategy consists on first computing an integral image (i.e. accumulated frequencies of visual word occurrences starting from an image corner, usually top-left) for every visual word in the vocabulary tree. Having the integral image pre-computed for all visual words, the histogram of visual word counts for an arbitrary sub-window can be computed with four operations instead of having to test if every detected feature falls inside the boundaries of the sub-window. Let $I_i$ be the integral image of a query image for node $i$ of the vocabulary tree, then the histogram $H$ of visual words counts for a given sub-window $W$ can be computed in the following way:

$$H_i = I_i(W_{br}) + I_i(W_{tl}) - I_i(W_{tr}) - I_i(W_{bl}) \qquad (11)$$

for all $i$, where $W_{br}$, $W_{tl}$, $W_{tr}$ and $W_{bl}$ are respectively the bottom right, top left, top right and bottom left coordinates of $W$.

The computational complexity of determining the visual word counts for an arbitrary sub-window is therefore $O(4 \cdot \varphi)$ operations, where $\varphi$ is the size of the vocabulary. Doing the same without integral images has a complexity of $O(5 \cdot \eta)$, where $\eta$ is the number of visual words found in the test image. From this, it is clear that integral images are a speed-up as long as $\varphi$ is significantly smaller than $\eta$ (e.g. in case of dense feature extraction from the image with a small vocabulary).

The second alternative is using a segmentation method to divide the image into a set of regions that must be recognized. Various options exist for this task which can be broadly classified as intensity based and, if stereo pairs of images are available, depth based. In this work we have evaluated one method of each type. Namely, an intensity based method similar to the watershed algorithm, and a depth based one.

## 4.1  Intensity-based Segmentation

The **intensity based** method we propose, that we called *floodcanny*, consists on first applying the Canny edge detector to the image, and use the resulting edges as hard boundaries in a *flood filling* segmentation process. In contrast with conventional watershed methods, in our method seed points are not local minima of the image, but are arbitrarily chosen from the set of unlabeled points; and a limit in brightness difference is imposed both for lower as well as for higher intensity values with respect to the seed point. For each candidate region of an acceptable size (in our experiments, having an area bigger than 900 pixels), a set of five sub-windows of different size centered in the segmented area are defined and evaluated. In general, it is intuitive to think that, the more accurate the segmentation of the image passed to the classifier is, the better will be the results of the object recognition method. More specifically, methods that can overcome highlights, shadows or weak reflections as the one proposed by [18] have a potential to provide more meaningful regions for the classifier, and the combination of such type of methods with appearance-based classifiers is an area of great interest, that we would address in future work. For the present work however, we have used only our proposed *floodcanny* method, which, despite of its simplicity, achieved good segmentation results as can be seen in Figure 9. Furthermore, it is fast to apply (less than 30 milliseconds for a $640 \times 480$ image), which is very convenient given our objectives.

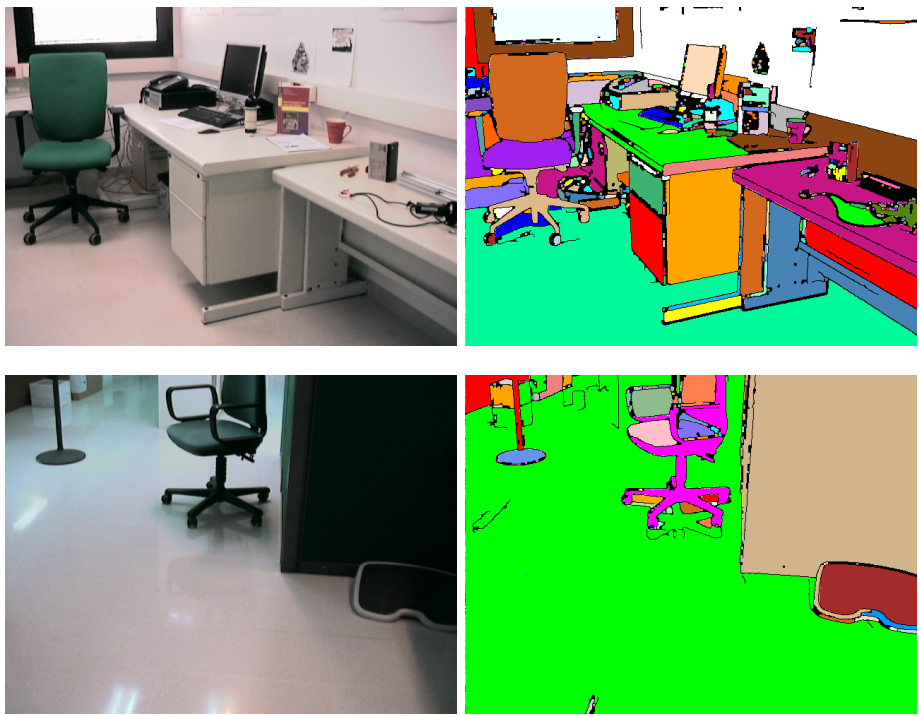

Figure 9: Results of the segmentation process using the *floodcanny* method. The first column shows the original images and the second column the segmented regions. Each color represents a different region, and Canny edges are superimposed for clarity.

## 4.2 Depth-based Segmentation

The second segmentation alternative proposed consisted of directly matching features between the left and right image to detect areas of constant depth. Since the geometry of the stereo cameras is known *a priori*, epipolar geometry constraints can be used together with the scale and orientation of a given feature to reduce the set of possible matches. To determine the possible location of the objects in the environment, a grid of 3D cells of different sizes is used. Reprojected features cast a vote for a cell of a grid if it lies within the 3D cell coordinates. Cells that have a minimum number of votes are reprojected to the image and added as a candidate window. It seems tempting to directly use the matched features to construct the histogram of feature word counts, as it would reduce the amount of background introduced in the visual word counts histogram. However, there is no guarantee that all features of the object have been detected in both images and matched, and the effects of missing important object features are potentially worse than introducing a small amount of background. Therefore we considered it more adequate to accept all visual words close to a set of valid matches.

## 4.3 Experimental Results

As in Section 3, an extensive cross-validation study has been conducted to evaluate the range of parameters of the method. For brevity here we only include the most relevant results and refer the interested reader to a technical report available online with all the experimental details[6]. This more detailed report includes experiments that address:

1. Floating point precision (single/double)

2. Histogram normalization method

3. Effect in computational time of inverted files

---

[6]http://www.iiia.csic.es/~aramisa/datasets/iiia30.html

4. Quality and number of training images

5. Different segmentation methods (i.e. sliding windows, intensity-based and depth-based segmetnation)

6. The effect of different widths and depths of the vocabulary tree

7. Number of nearest neighbors in the kNN classifier

8. Different types of feature detectors

9. Additional tests with manually pre-segmented image datasets.

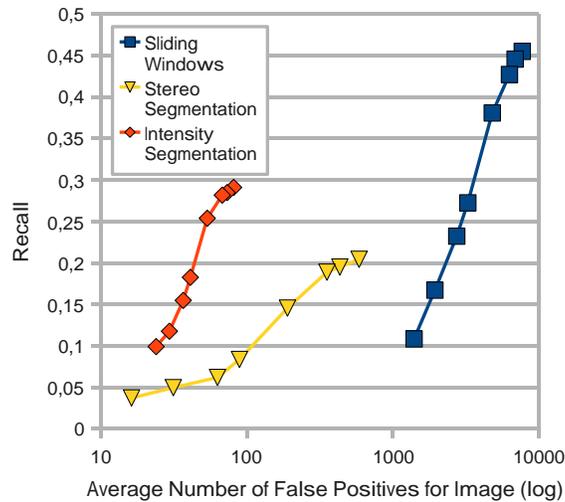

Figure 10: Results of applying Intensity Segmentation (the *floodcanny* algorithm), Stereo Segmentation and Sliding Windows to generate the sub-windows to evaluate at the first sequence of the IIIA30 dataset. For the three experiments the DoG detector and a tree with branch factor 10 and depth 4 have been used.

**Detection with Segmentation:** We have evaluated the proposed *floodcanny* intensity based segmentation algorithm and the depth based segmentation approach described earlier.

We applied the *floodcanny* to the first sequence of the IIIA30 dataset with good results. For each region sufficiently large, a set of five windows of different sizes, centered at the detected region is defined. Besides increasing recall, as can be seen in Figure 10, the number of false positives has decreased from thousands to only tens.

Despite this result, the proposed segmentation scheme is not optimal, as it usually works better for large and textureless objects, that can be segmented as a big single region. Contrarily, small and textured objects pose a problem to the *floodcanny* method, as no single large enough region can be found.

Regarding the depth segmentation, Figure 10 also shows the results for this experiment. Although the maximum attained recall is slightly lower than that of sliding windows, it must be noted that, at a similar level of recall, false positives are much lower.

## 4.4 Evaluation of Selected Configuration

In this Section we summarize the results obtained with the parameter configurations selected in the cross-validation study on all the test sequences.

Except for recall, which is better for the Vocabulary Tree method, the SIFT object recognition has better results in all other aspects related to robotics.

As can be seen in Table 6, with the segmentation schema adopted in this final experiment, we have obtained a recall better than with the SIFT method for untextured objects[7]. Unfortunately small and textured objects are harder to detect with the current segmentation , as they usually do not generate a large enough uniform region. However this is not a weakness of the Vocabulary Tree method but of the segmentation approach.

| Objects | 10nn | | 10nn with filtering $\delta = 0.8$ | | 5nn | | 1nn | | 10nn with relaxed overlap | |
|---|---|---|---|---|---|---|---|---|---|---|
| | Rec | Prec | Rec | Prec | Rec | Prec | Rec | Prec | Rec | Prec |
| Grey battery | 0.36 | 0.01 | 0.32 | 0.02 | 0.32 | 0.01 | 0.36 | 0.01 | 0.60 | 0.02 |
| Bicycle | 0.67 | 0 | 0.59 | 0 | 0.58 | 0.01 | 0.49 | 0.01 | 0.70 | 0 |
| Hartley book | 0.21 | 0 | 0.21 | 0 | 0.19 | 0 | 0.21 | 0 | 0.81 | 0.01 |
| Calendar | 0.18 | 0 | 0.09 | 0 | 0.15 | 0 | 0.12 | 0 | 0.53 | 0.01 |
| Chair 1 | 0.70 | 0.05 | 0.69 | 0.06 | 0.72 | 0.05 | 0.78 | 0.06 | 0.71 | 0.06 |
| Charger | 0.11 | 0 | 0 | 0 | 0 | 0 | 0 | 0 | 0.11 | 0 |
| Cube 2 | 0.11 | 0 | 0.11 | 0 | 0.11 | 0 | 0.17 | 0 | 0.28 | 0.01 |
| Monitor 3 | 0.77 | 0.16 | 0.77 | 0.17 | 0.66 | 0.14 | 0.71 | 0.09 | 0.93 | 0.21 |
| Poster spices | 0.46 | 0.02 | 0.46 | 0.02 | 0.35 | 0.02 | 0.46 | 0.03 | 0.59 | 0.03 |
| Rack | 0.60 | 0.06 | 0.58 | 0.07 | 0.60 | 0.07 | 0.58 | 0.06 | 0.82 | 0.09 |

Table 6: Precision and recall for some interesting objects of the IIIA30 dataset in the final Vocabulary Tree experiment (i.e. tree with branch factor 9 and depth 4, and features found with the Hessian Affine detector). Different choices of parameters for the classifier are displayed. Also, the last column, shows the results obtained using Equation 6 instead of Equation 5 to measure overlap.

Objects like the computer monitors, the chairs or the umbrella had a recall comparable to that of textured objects. As can be seen in Table 7, a similar recall was obtained for the objects of types textured and not textured. A slightly worse recall was obtained for the repetitively textured objects, but we believe it is mostly because of the segmentation method.

---
[7]For space reasons, only part of the Table was included. The full Table can be found in http://www.iiia.csic.es/~aramisa/datasets/iiia30_results

|        | 10nn | 10nn-0.8 | 5nn | 1nn | 10nn-relaxed |
|--------|------|----------|-----|-----|--------------|
| Repetitively textured objects | | | | | |
| Recall | 0.18 | 0.18 | 0.21 | 0.23 | 0.29 |
| Prec   | 0    | 0    | 0    | 0    | 0.01 |
| Textured objects | | | | | |
| Recall | 0.29 | 0.27 | 0.26 | 0.28 | 0.53 |
| Prec   | 0.02 | 0.02 | 0.02 | 0.02 | 0.02 |
| Not textured objects | | | | | |
| Recall | 0.29 | 0.26 | 0.27 | 0.29 | 0.39 |
| Prec   | 0.03 | 0.03 | 0.03 | 0.03 | 0.04 |

Table 7: Precision and recall depending on texture level of the objects in the final experiment with the [14] Vocabulary Tree. The objects are grouped in the same way as in Table 5. The title *10nn-0.8* stands for 10 nearest neighbors with filtering $\delta = 0.8$, and *10nn-relaxed* for 10 nearest neighbors with relaxed overlap.

Regarding the image quality parameters (see Table 8), the occluded objects obtained a higher recall level, but this was because, as mentioned in the previous discussion, the sliding windows approach taken in this experiment does not enforce a precise detection and, therefore, Equation 5 discards hypotheses correctly detecting object instances. When Equation 6 was used for all objects, instead of restricting it only to the occluded ones, recall for objects with *normal* and *blurred* viewing conditions is increased. The percentage of detected objects with a degree of overlap from 90% to 100% between the found and the ground truth bounding box was increased by 14%, showing that, although not precisely, the considered windows did overlap almost the whole object region.

|              | 10nn | 10nn-0.8 | 5nn  | 1nn  | 10nn-relaxed |
|--------------|------|----------|------|------|--------------|
| Normal       | 0.24 | 0.23     | 0.24 | 0.25 | 0.45 |
| Blur         | 0.29 | 0.28     | 0.28 | 0.3  | 0.46 |
| Occluded     | 0.64 | 0.61     | 0.62 | 0.62 | 0.64 |
| Illumination | 0.06 | 0.06     | 0.06 | 0.11 | 0.11 |
| Blur+Occl    | 0.43 | 0.41     | 0.43 | 0.46 | 0.43 |
| Occl+Illum   | 0.11 | 0.11     | 0.08 | 0.08 | 0.11 |
| Blur+Illum   | 0.14 | 0        | 0    | 0    | 0.14 |

Table 8: Recall depending on image characteristics. *Normal* stands for object instances with good image quality and *blur* for blurred images due to motion, *illumination* indicates that the object instance is in a highlight or shadow and therefore has low contrast. Finally the last three rows indicate that the object instance suffers from two different problems at the same time.

## 4.5 Discussion

With the selected configurations we obtained an average recall of 30%. More importantly, this approach has been able to detect objects that the SIFT could not find because of its restrictive matching stage. However, also 60 false positives per image on average were detected with the selected configuration, which represents a precision of 2% on average.

In the light of the performed experiments, it seems clear that the Vocabulary Tree method cannot be directly applied to a mobile robotics scenario, but some strategy to reduce the number of false positives is necessary. In addition to reducing false positives to acceptable levels, it is necessary to accelerate the detection step in order to process images coming from the robot cameras at an acceptable rate. Improving the segmentation strategy, or using a technique such as the one presented in [19] will surely help improve the accuracy.

Nevertheless, we found that the Vocabulary Tree method was able to detect objects that were inevitably missed by the SIFT Object Recognition method. Furthermore, new and promising *bag of features* type approaches are currently being proposed, such as the aforementioned [20] approach, the one by [21] and specially the one by [22]. In future work we plan to evaluate some of these methods.

Regarding the depth segmentation, Figure 10 also shows the results for this experiment. Although the maximum attained recall is slightly lower than that of sliding windows, it must be noted that, at a similar level of recall, false positives are much lower.

## 5  Viola-Jones Boosting

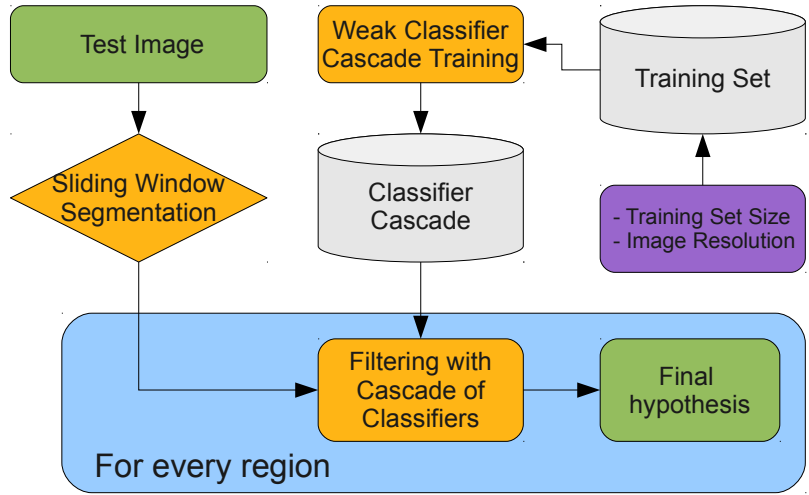

Figure 11: Diagram of the Viola and Jones Cascade of Weak Classifiers method, with tests shown as purple boxes. Orange boxes refer to steps of the method and green to input/output of the algorithm.

A third commonly used object recognition method is the cascade of weak classifiers proposed by Viola and Jones [17]. This method constructs a cascade of simple classifiers (i.e. simple Haar-like features in a certain position inside a bounding box) using a learning algorithm based on AdaBoost. Speed was of primary importance to the authors of [17], and therefore every step of the algorithm was designed with efficiency in mind. The method uses rectangular Haar-like features as input from the image, computed using Integral Images, which makes it a constant time operation regardless of the scale or type of feature. Then, a learning process that selects the most discriminative features constructs a cascade where each node is a filter that evaluates the presence of a single Haar-like feature with a given scale at a certain position in the selected region. The most discriminative filters are selected to be in the first stages of the cascade to discard windows not having the object of interest as soon as possible. At classification time, the image is explored using sliding windows. However, thanks to the cascade structure of the classifier it's only at interesting areas where processor time is really spent.

Notwithstanding its well known advantages, this approach suffers from significant limitations. The most important one being the amount of data required to train a competent classifier for a given class. Usually hundreds of positive and negative examples are required (e.g. in [23] 5000 positive examples, derived using random transformations from 1000 original training images, and 3000 negative examples where used for the task of frontal face recognition). Another known drawback is that a fixed aspect ratio of the objects is assumed with this method, that may not be constant for certain classes of objects (e.g. cars). Another drawback is the difficulty of generalizing the approach above 10 objects at a time[24]. Finally, the tolerance of the method to changes in the point of view is limited to about 20°. In spite of these limitations, the Viola and Jones object detector has had remarkable success and is widely used, especially for the tasks of car and frontal face detection.

Since the publication of the original work by Viola and Jones, many improvements to the method have appeared, for example to address the case of multi-view object recognition [25, 26].

## 5.1 Experimental Results

In this work the original method has been evaluated using a publicly available implementation[8]

**Training Set Size and Image Quality** As previously mentioned, one of the most important limitations of the Viola and Jones object recognition method is the size of the training set. In this work we have evaluated three different training sets. The first one consists of images extracted from the ground truth bounding boxes from test sequences IIIA30-2 and IIIA30-3. The second one consists of the same training set used for the Vocabulary Tree experiments (20 good quality training images per object type) and additional synthetic views generated from these images. Finally, the third training set is a mix between good quality images extracted from videos recorded with a digital camera (for 21 objects, between 700 and 1200 manually segmented images per object), and a single training image plus 1000 new synthetic views (for 8 objects).

The dataset used for the first test only had a few images for each type of object: 50 to 70 images per class. In Table 9 the results obtained for sequences IIIA30-1 and IIIA30-2 are shown. With so few training data, the Viola and Jones classifier is able to find only some instances for objects of 11 out of the 29 categories. This performance is expected due to the limited amount of training data.

Table 10 shows the results obtained with the twenty training images used in the Vocabulary Tree experiments, but further enhancing the set by synthetically generating a hundred extra images for each training sample. As it can be seen, the usage of high quality images and the synthetic views significantly improved the results.

Finally, Table 11 shows the results obtained using the third training set, which consisted of hundreds of good quality images extracted from video record-ings done with a conventional camera. A conclusion that can be quickly inferred from the table is the decrease in performance caused by occlusions. Even objects that achieve a good recall and precision with good viewing conditions, fail in the case of occlusions. In contrast, blurring and illumination variations did not affect performance significantly. Regarding the object types, (textured, untextured and repetitively textured) textured objects obtained an overall recall of 26% and precision of 33%, similar to that of repetitively textured objects (24% recall and 36% precision). Finally, untextured objects obtained 14% of recall and 19% precision. With this dataset, the average f-measure obtained is higher than the one obtained with the bag of features object detection method.

---

[8]We have used the implementation that comes with the OpenCV 1.0 library: http://sourceforge.net/projects/opencvlibrary/

| Object | Recall | Prec | Object | Recall | Prec |
|---:|:---:|:---:|---:|:---:|:---:|
| Grey battery | 0.0 | 0.0 | Monitor 2 | 0.14 | 0.14 |
| Red battery | 0.28 | 0.02 | Monitor 3 | 0.03 | 0.01 |
| Bicycle | 0.46 | 0.07 | Orbit box | 0.03 | 0.01 |
| Ponce book | 0.0 | 0.0 | Dentifrice | 0.0 | 0.0 |
| Hartley book | 0.03 | 0.01 | Poster CMPI | 0.17 | 0.15 |
| Calendar | 0.19 | 0.01 | Phone | 0.0 | 0.0 |
| Chair 1 | 0.11 | 0.22 | Poster Mystrands | 0.36 | 0.27 |
| Chair 2 | 0.71 | 0.05 | Poster spices | 0.46 | 0.06 |
| Chair 3 | 0.0 | 0.0 | Rack | 0.0 | 0.0 |
| Charger | 0.0 | 0.0 | Red cup | 0.0 | 0.0 |
| Cube 1 | 0.0 | 0.0 | Stapler | 0.03 | 0.01 |
| Cube 2 | 0.0 | 0.0 | Umbrella | 0.03 | 0.02 |
| Cube 3 | 0.0 | 0.0 | Window | 0.36 | 0.2 |
| Extinguisher | 0.0 | 0.0 | Wine bottle | 0.0 | 0.0 |
| Monitor 1 | 0.0 | 0.0 | | | |

Table 9: Recall and precision values obtained training the Viola & Jones object detector using images extracted from the IIIA30-3 sequence and evaluating in sequences IIIA30-1 and IIIA30-2.

| Object | Recall | Prec | Object | Recall | Prec |
|---:|:---:|:---:|---:|:---:|:---:|
| Grey battery | 0.01 | 0.02 | Monitor 2 | 0.41 | 0.20 |
| Red battery | 0.08 | 0.04 | Monitor 3 | 0.40 | 0.18 |
| Bicycle | 0.01 | 0.10 | Orbit box | 0.10 | 0.16 |
| Ponce book | 0.08 | 0.31 | Dentifrice | 0.01 | 0.03 |
| Hartley book | 0.04 | 0.08 | Poster CMPI | 0.10 | 0.05 |
| Calendar | 0.11 | 0.27 | Phone | 0.07 | 0.08 |
| Chair 1 | 0.02 | 0.30 | Poster Mystrands | 0.71 | 0.12 |
| Chair 2 | 0.01 | 0.34 | Poster spices | 0.05 | 0.05 |
| Chair 3 | 0.02 | 0.05 | Rack | 0.06 | 0.55 |
| Charger | 0.0 | 0.08 | Red cup | 0.01 | 0.05 |
| Cube 1 | 0.06 | 0.21 | Stapler | 0.02 | 0.20 |
| Cube 2 | 0.0 | 0.56 | Umbrella | 0.05 | 0.58 |
| Cube 3 | 0.03 | 0.24 | Window | 0.10 | 0.08 |
| Extinguisher | 0.09 | 0.13 | Wine bottle | 0.03 | 0.32 |
| Monitor 1 | 0.02 | 0.01 | | | |

Table 10: Recall and precision values for each object category for the Viola & Jones object detector when using the same training set as with the bag of features with synthetically generated images.

The performance on the posters is surprisingly low in comparison to the other two methods. The explanation could be the large changes in point of view that the posters suffer through the video sequences. The time necessary

|  | All | | Non-Occluded | | Occluded | |
| --- | --- | --- | --- | --- | --- | --- |
| Object | Recall | Prec | Recall | Prec | Recall | Prec |
| Grey battery | 0.36 | 0.24 | 0.41 | 0.24 | 0.0 | 0.0 |
| Red battery | 0.37 | 0.82 | 0.44 | 0.82 | 0.0 | 0.0 |
| Bicicle | 0.0 | 0.0 | 0.0 | 0.0 | 0.0 | 0.0 |
| Ponce book | 0.81 | 0.88 | 0.86 | 0.86 | 0.25 | 0.02 |
| Hartley book | 0.66 | 0.94 | 0.70 | 0.94 | 0.0 | 0.0 |
| Calendar* | 0.33 | 0.08 | 0.38 | 0.08 | 0.0 | 0.0 |
| Chair 1 | 0.0 | 0.0 | 0.0 | 0.0 | 0.0 | 0.0 |
| Chair 2* | 0.0 | 0.0 | 0.0 | 0.0 | 0.0 | 0.0 |
| Chair 3 | 0.0 | 0.0 | 0.0 | 0.0 | 0.0 | 0.0 |
| Charger | 0.12 | 0.08 | 0.12 | 0.08 | 0.0 | 0.0 |
| Cube 1 | 0.22 | 0.43 | 0.23 | 0.29 | 0.2 | 0.15 |
| Cube 2 | 0.23 | 0.11 | 0.20 | 0.09 | 0.34 | 0.03 |
| Cube 3 | 0.28 | 0.53 | 0.37 | 0.48 | 0.09 | 0.06 |
| Extinguisher | 0.0 | 0.0 | 0.0 | 0.0 | 0.0 | 0.0 |
| Monitor 1* | 0.0 | 0.0 | 0.0 | 0.0 | 0.0 | 0.0 |
| Monitor 2* | 0.23 | 0.57 | 0.39 | 0.57 | 0.0 | 0.0 |
| Monitor 3* | 0.04 | 0.13 | 0.05 | 0.13 | 0.0 | 0.0 |
| Orbit box* | 0.15 | 0.03 | 0.17 | 0.03 | 0.0 | 0.0 |
| Dentifrice | 0.0 | 0.0 | 0.0 | 0.0 | 0.0 | 0.0 |
| Poster CMPI | 0.11 | 0.34 | 0.19 | 0.34 | 0.0 | 0.0 |
| Phone | 0.05 | 0.09 | 0.0 | 0.0 | 0.3 | 0.09 |
| Poster Mystrands | 0.0 | 0.0 | 0.0 | 0.0 | 0.0 | 0.0 |
| Poster spices | 0.04 | 0.38 | 0.12 | 0.38 | 0.0 | 0.0 |
| Rack | 0.0 | 0.0 | 0.0 | 0.0 | 0.0 | 0.0 |
| Red cup | 0.89 | 0.89 | 0.89 | 0.89 | 0.0 | 0.0 |
| Stapler | 0.24 | 0.21 | 0.24 | 0.21 | 0.0 | 0.0 |
| Umbrella | 0.0 | 0.0 | 0.0 | 0.0 | 0.0 | 0.0 |
| Window | 0.03 | 0.40 | 0.10 | 0.40 | 0.0 | 0.0 |
| Wine bottle* | 0.10 | 0.06 | 0.10 | 0.06 | 0.0 | 0.0 |

Table 11: Recall and precision values for each object category using the Viola & Jones object detector. When we decompose the precision-recall values for occluded and non-occluded objects, results shows a performance drop for occluded objects. The asterisk mark denotes objects trained from synthetic images.

to apply the classifiers for all the classes to one test image is 728 ms on average.

### 5.1.1 Discussion

Despite the use of very simple image features, the Viola and Jones Cascade of classifiers attains a good level of precision and recall for most of the objects in a very low runtime. Its main drawbacks are the large, in comparison with the other evaluated techniques, training dataset required to obtain a good level of performance, and the limited robustness to changes in the point of view and occlusions of the method. Furthermore, some theoretically "easy" objects, such as the posters, proved to be troublesome to the Viola and Jones method. This is probably due to overfitting to some particular view, or to too much variability of the very rich Haar feature distribution when changing the point of view, where the method was unable to find any recognizable regular pattern.

Nevertheless, the idea of a boosted cascade of weak classifiers is not limited to the very fast but simple Haar features, but any kind of classifier can be used for that matter. A very interesting alternative is using linear SVMs as weak classifiers, since it allows to add a non-linear layer to an already efficient linear classifier. Such idea has been already successfully applied in a few cases [27, 28], and we believe it is a very interesting line to investigate.

# 6  Conclusions

Object perception capabilities are a key element in building robots able to develop useful tasks in generic, unprepared, human environments. Unfortunately, state of the art papers in computer vision do not evaluate the algorithms with the problems faced in mobile robotics. In this work we have contributed an evaluation of three object recognition algorithms in the difficult problem of object recognition in a mobile robot: the SIFT object recognition method, the Vocabulary Tree and a boosted cascade of weak classifiers. In contrast with the case of high-quality static Flickr photos, images acquired by a moving robot are likely to be low resolution, unfocused and affected by problems like bad framing, motion blur or inadecuate illumination, due to the short dynamic range of the camera. The three methods have been thoroughly evaluated in a dataset obtained by our mobile robot while navigating in an unprepared indoor environment. Finally, in order to improve the performance of the methods, we have also proposed several improvements to the methods.

This work aims to be a practical help for roboticists that want to enable their mobile robots with visual object recognition capabilities, highlighting the advantages and drawbacks of each method and commenting on its applicability in practical scenarios. Furthermore, relevant enhancements for the methods existent in the literature (i.e. support for 3D models in the SIFT object recognition method) are reported.

We have created a challenging dataset of video sequences with our mobile robot while moving in an office type environment. These sequences have been acquired at a resolution of $640 \times 480$ pixels with the robot cameras, and are full of blurred images due to motion, large viewpoint and scale changes and object occlusions.

The first evaluated method is the SIFT object recognition method, proposed by [11]. Many issues including:

- training image quality
- approximate local descriptor matching
- false hypotheses filtering methods

are evaluated in a subset of the proposed dataset. Furthermore, we propose and evaluate several modifications to the original schema to increase the detected objects and reduce the computational time.

The parameter settings that attained best overall results are subsequently tested in the rest of the dataset and carefully evaluated to have a clear picture of the response that can be expected from the method with respect to untextured objects or image degradations. Next, a similar evaluation is carried on for the second method, the Vocabulary Tree proposed by [14]. For the case of the Viola and Jones cascade of weak classifiers the used implementation directly offers a thoroughly evaluated selection of parameters, and the main variable we have evaluated is the training set size.

From the results obtained, it can be seen that with the present implementation of the methods, the SIFT object recognition method adapts better to the performance requirements of a robotics application. Furthermore, it is easy to train, since a single good quality image sufficed to attain good recall and precision levels. However, although this method is resistant to occlusion and reasonable levels of motion blur, its usage is mostly restricted to flat well textured objects. Also, classification (generalizing to unseen object instances of the same class) is not possible with this approach.

On the other hand, the Vocabulary Tree method has obtained good recognition rates both for textured and untextured objects, but too many false positives per image were found. Finally, the Viola and Jones method offers both a good recall (specially for low-textured objects) and execution speed, but is very sensitive to occlusions and the simple features used seem to be unable to cope with the most richly textured objects in case of strong changes in point of view.

Although we have evaluated the proposed object recognition methods in a wide range of dimensions, one that is lacking is a more in-depth study of how the composition and size of the training set affects the overall results. For example, having similar objects, as the different monitors or chairs in the IIIA30 dataset, can cause confusion to the methods. Therefore future work will address the evaluation of different sub-sets of target objects.

The main limitation of the SIFT object recognition method is that only the first nearest neighbor of each test image feature is considered in the subsequent stages. This restriction makes the SIFT method very fast, but at the same time makes it unable to detect objects with repetitive textures. Other approaches with direct matching, like that of [29], overcome this by allowing every feature to vote for all feasible object hypotheses given the feature position and orientation. Evaluating this type of methods, or modifying the SIFT to accept several hypotheses for each test image feature, would be an interesting line of continuation of this work.

The sliding windows approach could be improved by allowing windows with a good probability of a correct detection to inhibit neighboring and/or overlapping windows, or simply keeping the best window for a given object would clearly reduce the number of false positives.

Regarding the segmentation schema, we believe that results can be improved by adopting more reliable techniques, able to resist highlights and shadows. Besides, textured areas pose a problem to the segmentation algorithm as, with the current technique, no windows will be cast in scattered areas. It would be interesting to test if a Monte Carlo approach to fuse neighboring regions can help alleviate the problem without significantly affecting the computational time. Also a voting mechanism to detect areas with a high number of small regions can be attempted.

The Viola and Jones approach was the fastest of the three in execution time and, as mentioned earlier, it obtained a reasonable level of precision and recall –especially for the low-textured objects–, but at the cost of a significantly larger training effort –both in computational cost and labeled data– than the other two methods. In addition, objects instances with occlusions had a performance notably lower in comparison.

More powerful features, like the ones used for the other two methods, or the popular HOGs [30], could also be used in the Viola and Jones cascade of classifiers. However that would increase the computational cost of the method. In order to handle the viewpoint changes extensions have been proposed to the method [25, 31], specially using Error-Correcting Output Codes (ECOC) [32]. It would be interesting to evaluate the impact on the performance of these extensions.

In summary: Three fundamentally different methods, each one a representative of a very successful established paradigm for visual object perception, have been evaluated for feasibility for the particular task of object detection in a mobile robot platform. Furthermore, a number of variations or improvements to the selected methods are being actively produced and evaluated.

# Acknowledgements

This work was supported by the FI grant from the Generalitat de Catalunya, the European Social Fund, and the MID-CBR project grant TIN2006-15140-C03-01 and FEDER funds and the grant 2005-SGR-00093 and theMIPRCV Consolider Imagennio 2010 and by the Rio Tinto Centre for Mine Automation and the ARC Centre of Excellence programme, funded by the Australian Research Council (ARC) and the New South Wales State Government.